%% file: main.tex
\documentclass[10pt,twocolumn,letterpaper]{article}

\usepackage[pagenumbers]{cvpr} 

\newtheorem{remark}{Remark}
\usepackage{algorithm}
\usepackage{algorithmic}
\usepackage[accsupp]{axessibility}


\definecolor{cvprblue}{rgb}{0.21,0.49,0.74}
\usepackage[pagebackref,breaklinks,colorlinks,allcolors=cvprblue]{hyperref}

\def\paperID{4223} 
\def\confName{CVPR}
\def\confYear{2025}

\title{Rethinking Epistemic and Aleatoric Uncertainty for Active Open-Set Annotation: An Energy-Based Approach}

\author{Chen-Chen Zong, Sheng-Jun Huang\thanks{Corresponding Author} \\
Nanjing University of Aeronautics and Astronautics\\
Nanjing, 211106, China\\
{\tt\small \{chencz,huangsj\}@nuaa.edu.cn}
}

\begin{document}
\maketitle
\input{sec/0_abstract}    
\input{sec/1_intro}

\input{sec/2_relate}

\input{sec/3_method}
\input{sec/4_experi}

\input{sec/5_conclu}

\section*{Acknowledgements}

This work was supported by the Natural Science Foundation of Jiangsu Province of China (BK20222012) and the NSFC(U2441285, 62222605).

{
    \small
    \bibliographystyle{ieeenat_fullname}
    \bibliography{main}
}
\appendix
\input{sec/X_suppl}


\end{document}

%% file: sec/0_abstract.tex
\begin{abstract}
Active learning (AL), which iteratively queries the most informative examples from a large pool of unlabeled candidates for model training, faces significant challenges in the presence of open-set classes. Existing methods either prioritize query examples likely to belong to known classes, indicating low epistemic uncertainty (EU), or focus on querying those with highly uncertain predictions, reflecting high aleatoric uncertainty (AU). However, they both yield suboptimal performance, as low EU corresponds to limited useful information, and closed-set AU metrics for unknown class examples are less meaningful. In this paper, we propose an Energy-based Active Open-set Annotation (EAOA) framework, which effectively integrates EU and AU to achieve superior performance. EAOA features a $(C+1)$-class detector and a target classifier, incorporating an energy-based EU measure and a margin-based energy loss designed for the detector, alongside an energy-based AU measure for the target classifier. Another crucial component is the target-driven adaptive sampling strategy. It first forms a smaller candidate set with low EU scores to ensure closed-set properties, making AU metrics meaningful. Subsequently, examples with high AU scores are queried to form the final query set, with the candidate set size adjusted adaptively. Extensive experiments show that EAOA achieves state-of-the-art performance while maintaining high query precision and low training overhead. The code is available at this \href{https://github.com/chenchenzong/EAOA}{link}.
\end{abstract}

%% file: sec/1_intro.tex
\section{Introduction}
\label{sec:intro}

The success of deep neural networks (DNNs) is largely fueled by large-scale, accurately labeled datasets \cite{lecun2015deep,zong2024dirichlet}. However, acquiring such data is often expensive and time-consuming primarily due to the labor-intensive nature of manual labeling \cite{ren2021survey,10.1007/978-3-031-73390-1_8}. To save labeling costs, recent research has focused on developing effective model training techniques that leverage limited and insufficiently labeled data \citep{zhou2018brief,van2020survey,ren2021survey}. Among them, active learning (AL) has emerged as a popular framework, which iteratively selects the most informative examples from the unlabeled data pool and queries their labels from the Oracle \cite{settles2009active,ren2021survey, huang2021asynchronous}.

\begin{figure}[t]
  \centering
   \includegraphics[width=0.97\linewidth]{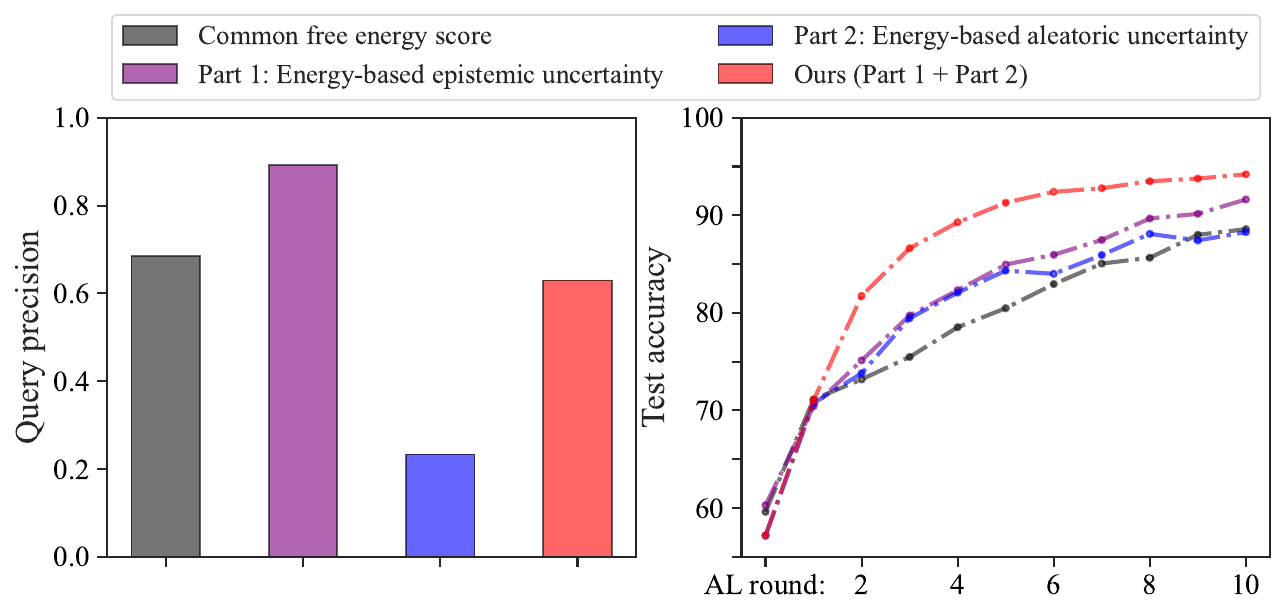}
   \caption{Dataset: CIFAR-10; mismatch ratio: 40\%. Our motivation: in open-set scenarios, querying examples with low epistemic uncertainty yields high query precision, but the overall information content is low, resulting in poor model performance. Focusing on examples with high aleatoric uncertainty also leads to poor performance, as the model's assessment of this uncertainty becomes meaningless for open-set examples. Nevertheless, an effective combination of the two can lead to a superior outcome.}
   \label{fig:motivation}
\end{figure}

Existing AL methods can be categorized into three types based on their sampling strategies: uncertainty-based \cite{balcan2007margin,holub2008entropy,yoo2019learning,10.1007/978-3-031-73390-1_8}, diversity-based \cite{nguyen2004active,sener2017active,xie2023active}, and hybrid strategies \cite{huang2010active,ash2019deep,safaei2024entropic}. Most of these methods operate under the closed set assumption, which posits that the classes in the unlabeled data match those in the labeled data. However, in real-world scenarios, it is challenging and often costly to ensure that no open-set classes are present in the unlabeled data. Meanwhile, several studies \cite{du2021contrastive,ning2022active,park2022meta,safaei2024entropic} have shown that these methods perform poorly when open-set classes are involved, as such examples tend to receive uncertain model predictions and exhibit distinct features. Therefore, developing effective AL methods for open-world scenarios, where open-set classes exist, is of significant importance.

Recently, this emerging research problem has garnered considerable attention \cite{du2021contrastive,ning2022active,park2022meta,safaei2024entropic,10.1007/978-3-031-73390-1_8}. For instance, in LfOSA \cite{ning2022active}, the authors formulate this problem as open-set annotation (OSA) and utilize queried unknown class examples to train a $(C+1)$-class detector for rejecting open-set examples and focusing more on selecting known class ones. Two recent methods, EOAL \cite{safaei2024entropic} and BUAL \cite{10.1007/978-3-031-73390-1_8}, adopt a structure similar to LfOSA, with EOAL aimed at improving the recognition of known class examples, while BUAL focuses more on sampling highly uncertain examples. Despite demonstrating high query precision, our findings reveal that these methods can not achieve satisfactory test accuracy and struggle to identify the most informative examples.

To analyze the reasons behind their failure, we review the concept of uncertainty quantification\footnote{A detailed introduction with intuitive illustrations is in \cref{sec:uq}. }. Epistemic uncertainty refers to a measure that remains high for instances not previously encountered and decreases when these instances are included in the training \cite{kendall2017uncertainties,smith2018understanding}. In contrast, aleatoric uncertainty is characterized by elevated values in ambiguous examples \cite{kendall2017uncertainties,smith2018understanding}. In closed-set settings, examples with high epistemic uncertainty tend to reside in low-density regions of the representation space, while those with high aleatoric uncertainty may appear around the decision boundary due to ambiguous features. Both types are potential targets for our queries. However, in open-set scenarios, examples with high epistemic uncertainty are likely to be open-set instances, and aleatoric uncertainty is meaningful only in closed-set contexts, as it reflects the ambiguity between different observable classes\footnote{Essentially, the class probability $p(y|x)=\frac{p(x,y)}{p(x)} $ is meaningful only when $p(x)\ne 0$, i.e., $x$ must be an observing example from a known class.} \cite{kendall2017uncertainties,mukhoti2023deep}. Therefore, selecting examples with low epistemic uncertainty (to ensure a closed set) and high aleatoric uncertainty is a more reasonable choice. However, LfOSA and EOAL focus on querying examples with low epistemic uncertainty, while BUAL prioritizes querying those with high aleatoric uncertainty. This may be the potential reason for their failure.



To validate this, we present the results in Figure \ref{fig:motivation}. As shown, focusing solely on either epistemic or aleatoric uncertainty in an open-world scenario results in suboptimal performance. However, effectively combining both leads to a significant improvement. Inspired by this, we propose \textbf{E}nergy-based \textbf{A}ctive \textbf{O}pen-set \textbf{A}nnotation (EAOA), an approach that effectively queries the most informative examples by considering both types of uncertainty. EAOA maintains two networks: a $(C+1)$-class detector and a $C$-class target classifier, with the following contributions:
\begin{itemize}
    \item An energy-based epistemic uncertainty measure is designed for the detector, expressed as the free energy score on known classes minus that on the unknown class. This measure integrates both learning-based and data-driven perspectives, enabling reliable uncertainty assessment in data-limited scenarios.

    \item An energy-based aleatoric uncertainty measure is proposed for the target classifier, defined as the free energy score on all classes minus that on secondary classes.

    \item A coarse-to-fine querying strategy is proposed. It first selects examples with low epistemic uncertainty to form a smaller candidate set, ensuring closed-set properties, which makes aleatoric uncertainty meaningful. Then, it queries examples with high aleatoric uncertainty within this set, with the candidate set size adaptively adjusted through a novel target-driven strategy.

    \item A margin-based energy loss is introduced for the detector training, aimed at maximizing the free energy score on known classes while minimizing that on the unknown class, thereby enhancing the unknown class detection.

    \item Extensive experiments show that EAOA outperforms current state-of-the-art methods in test accuracy, query precision, and training efficiency.

\end{itemize}

%% file: sec/2_relate.tex
\section{Related Work}
\label{sec:related}

\textbf{Active learning (AL)} has garnered great research interest as a primary framework for reducing labeling costs by querying the most informative examples for model training. AL's query methods can be categorized based on their data sources into three types: query-synthesizing \cite{mahapatra2018efficient,mayer2020adversarial,zhu2017generative}, stream-based \cite{fang2017learning,narr2016stream}, and pool-based approaches \cite{balcan2007margin,holub2008entropy,huang2021asynchronous,du2021contrastive,10.1007/978-3-031-73390-1_8}. Among these, pool-based methods are currently the mainstream, operating under the assumption of a large pool of available unlabeled data, from which a subset of examples is selected for annotation in each AL round. These query methods can be further divided into three categories: 1) uncertainty-based strategies \cite{li2006confidence,balcan2007margin,holub2008entropy,yoo2019learning}, which select instances for which labeling is least certain; 2) diversity-based strategies \cite{nguyen2004active,sener2017active,xie2023active}, which query instances that are most representative or exhibit the greatest feature diversity; and 3) hybrid strategies \cite{huang2010active,ash2019deep,safaei2024entropic}, which combine both to achieve better performance.

\textbf{Open-set recognition (OSR)} refers to a system's ability to differentiate between data types it has encountered during training (in-distribution (ID) data) and those it has not previously seen (out-of-distribution (OOD) data). Earlier studies employed traditional machine learning techniques such as support vector machines \cite{jain2014multi,scheirer2014probability}, extreme value theory (EVT) \cite{zhang2016sparse}, nearest class mean classifier \cite{bendale2015towards}, and nearest neighbor \cite{mendes2017nearest}. Recently, there has been growing interest in using generative models to learn representation spaces focused exclusively on known examples \cite{oza2019c2ae,sun2020conditional,zhang2020hybrid,perera2020generative}. Other techniques often aim to simulate unknown examples, providing a more intuitive approach to OSR \cite{ge2017generative,neal2018open,chen2020learning,zhou2021learning,chen2021adversarial}. However, simply applying these methods in AL scenarios under the open-world assumption often leads to failure for two main reasons. First, the recognition performance of these methods heavily relies on the classifier's effectiveness; when the classifier underperforms—a common occurrence in AL scenarios—the overall performance can decline significantly \cite{vaze2021open}. Second, some genuine unknown class examples are inevitably labeled during the labeling process, and OSR methods may not effectively utilize them.

\begin{figure*}[t]
  \centering
   \includegraphics[width=0.99\linewidth]{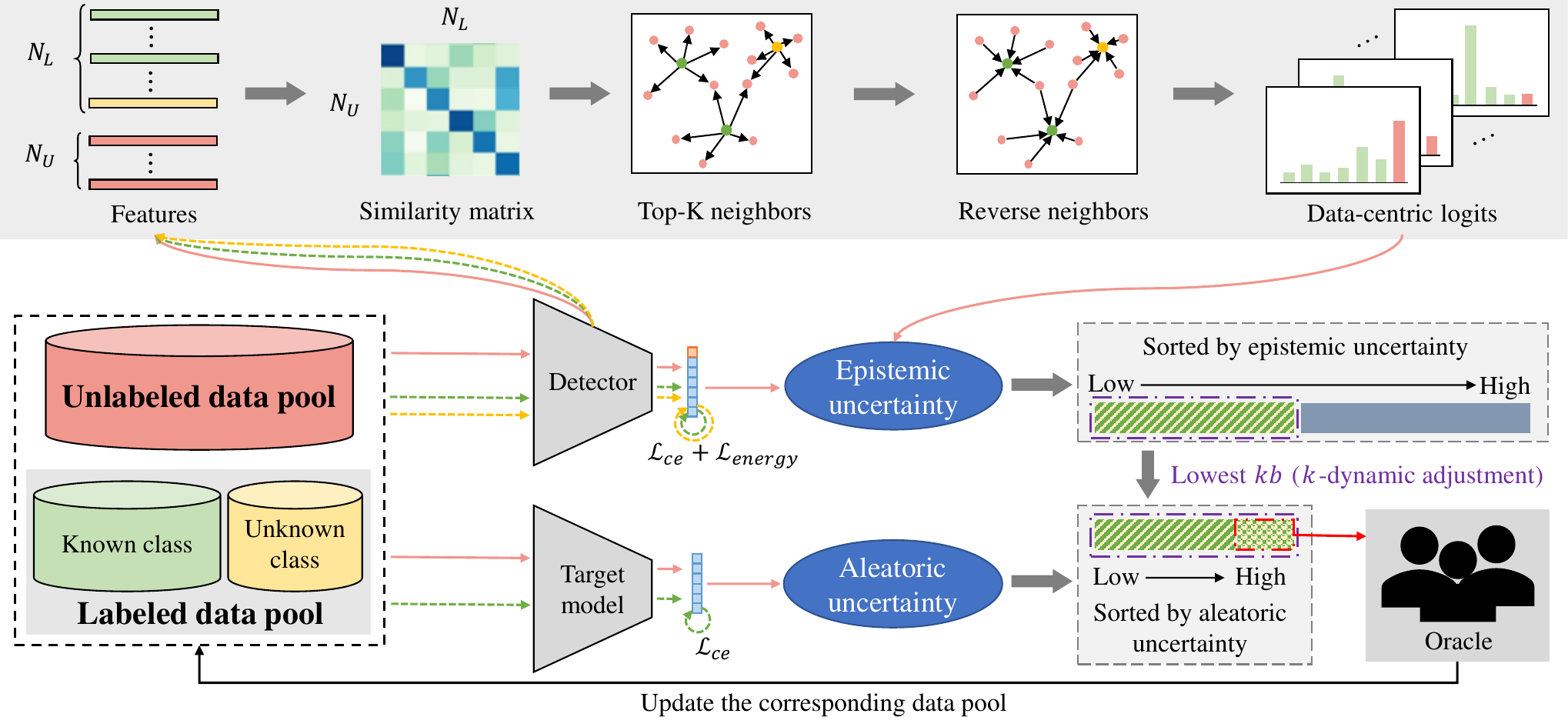}

   \caption{The framework of EAOA. It consists of three general steps: model training, example selection, and Oracle labeling. In the model training phase, a detector is trained to assess epistemic uncertainty (EU) from both learning-based and data-driven perspectives, along with a target classifier to evaluate aleatoric uncertainty (AU) based on class confusion. In the example selection phase, $kb$ examples with the lowest EU scores are chosen first, followed by querying $b$ examples with the highest AU scores, where $k$ is adaptively adjusted based on the target precision. In the Oracle labeling phase, the queried examples are assigned labels, and all relevant data pools are updated accordingly.}
   \label{fig:framework}
\end{figure*}

\textbf{Active open-set annotation (AOSA)} refers to AL tasks under open-world scenarios, which aligns more closely with practical application scenarios and has become a research hotspot in recent years \cite{du2021contrastive,ning2022active,park2022meta,safaei2024entropic,10.1007/978-3-031-73390-1_8}. CCAL \cite{du2021contrastive} and MQNet \cite{park2022meta} employ contrastive learning and established metrics respectively to assess sample purity and informativeness, utilizing heuristic and meta-learning approaches, respectively, to achieve a balance. However, by not fully leveraging labeled unknown class examples, they provide inadequate assessments, leading to poorer model performance. LfOSA \cite{ning2022active} incorporates labeled unknown class examples to train an additional $(C+1)$-class classifier (a.k.a, detector), using the maximum activation values (MAVs) to identify known class examples. EOAL \cite{safaei2024entropic} enhances the detector by adding an additional binary classifier head and uses entropy values, calculated separately for known and unknown classes, to identify known class examples. BUAL \cite{10.1007/978-3-031-73390-1_8} defines positive uncertainty and negative uncertainty respectively, and utilizes the detector's OOD probabilities to balance, aiming to identify highly uncertain examples. However, as previously noted, these methods fail to select examples with both low epistemic uncertainty and high aleatoric uncertainty, resulting in suboptimal performance.

%% file: sec/3_method.tex
\section{Methodology}
\label{sec:method}

\subsection{Preliminaries}

\textbf{Notations.} Consider the problem of ordinary $C$-class classification. In active open-set annotation (AOSA) tasks, we start with a limited labeled dataset $\mathcal{D}_{L}^{kno}=\left \{ \left ( x_i^L,y_i^L \right )  \right \} _{i=1}^{\mathcal{N} _L}$ containing $\mathcal{N} _L$ examples from known classes for training, alongside a sufficiently large unlabeled data pool $\mathcal{D}_{U}=\left \{ x_i^U \right \} _{i=1}^{\mathcal{N} _U}$ consisting of $\mathcal{N} _U$ examples from both known and unknown classes for querying. The label $y_i^L$ of an instance $x_i^L$ belongs to $\left \{ 1,\cdots ,C \right \}$, whereas the label $y_i^U$ of an instance $x_i^U$ is not provided prior to Oracle labeling and falls within $\left \{ 1,\cdots, C+1 \right \}$, with $C+1$ representing all unknown classes. At each active learning (AL) cycle, a batch of $b$ examples, denoted as $X^{query}=X_{kno}^{query}\cup X_{unk}^{query}$, is selected according to a specified query strategy and sent to Oracle for labeling. Then, $\mathcal{D}_{L}^{kno}$ and $\mathcal{D}_{U}$ are updated accordingly, and $X_{unk}^{query}$ forms the unknown class dataset $\mathcal{D}_{L}^{unk}$ with $\mathcal{D}_{L} = \mathcal{D}_{L}^{kno}\cup \mathcal{D}_{L}^{unk}$.

\textbf{Overview.} As outlined in the Introduction, addressing the AOSA problem requires first ensuring that the queried examples exhibit low epistemic uncertainty to approximate a closed set, enabling meaningful aleatoric uncertainty assessment, and then querying examples with high aleatoric uncertainty.
To achieve this goal, we propose an \textbf{E}nergy-based \textbf{A}ctive \textbf{O}pen-set \textbf{A}nnotation (EAOA) framework, as illustrated in Figure \ref{fig:framework}, which primarily consists of three key components: active sampling, detector training, and classifier training. Specifically, we begin by training a detector network to evaluate the epistemic uncertainty of examples from both learning-based and data-driven perspectives, leveraging the labeled data from both known and unknown classes. Next, we assess the aleatoric uncertainty of examples by utilizing the free energy discrepancy in predictions made by the target classifier. Finally, we query a smaller candidate set of examples with low epistemic uncertainty first, with the set dynamically adjusting by rounds, and then acquire the query set with high aleatoric uncertainty.

\subsection{Energy-based Epistemic and Aleatoric Uncertainty for Active Sampling}

\textbf{Energy-based epistemic uncertainty.}
Energy-based models (EBMs) \cite{liu2020energy,wang2021can} build a function $E\left ( x \right ):\mathbb{R}^D \to \mathbb{R}$ to map a $D$-dim data point to a scalar and defines the probability distribution in multi-class settings through logits as:
\begin{equation}
\label{eq1}
    p\left ( y|x \right ) =\frac{e ^{-E\left ( x,y \right ) }}{\int_{y}e^{-E\left ( x,y \right ) } }=\frac{e ^{f_y\left ( x \right ) }}{\sum_{c=1}^{C} e^{f_c\left ( x \right ) } } =\frac{e ^{-E\left ( x,y \right ) }}{e ^{-E\left ( x\right ) }}   ,
\end{equation}
where $f_y(x)$ denotes the predicted logit of model $f$ for instance $x$ regarding label $y$, $E(x,y)=-f_y(x)$, and $E\left ( x\right ) = -\log_{}{\sum_{c=1}^{C}e^{-E\left ( x,c \right ) } } $ is called free energy. The probability density of $x$ in an EBM can be written as:
\begin{equation}
\label{eq2}
    p\left ( x \right ) =\frac{e^{ -E\left ( x \right )}  }{ \int_{x}  e^{ -E\left ( x \right )} }=\frac{\int_{y}e^{-E\left ( x,y \right ) }}{\int_{x}\int_{y}e^{-E\left ( x,y \right ) }} =\frac{e^{ -E\left ( x \right )} }{\mathcal{Z}   } .
\end{equation}
This implies that for two data points, $x_1$ and $x_2$, if $E\left ( x_1 \right ) > E\left ( x_2 \right )$, then $x_1$ lies in a sparser region compared to $x_2$ w.r.t. the marginal distribution. This aligns with epistemic uncertainty: high-uncertainty examples are distributed in low-density regions due to their lower occurrence frequency. 

Nevertheless, free energy is not ideal for directly measuring epistemic uncertainty in the AOSA task, as the underlying EBM does not fully utilize the information contained in the labeled unknown class examples. To counter this, we group all unknown classes into the $C+1$ category, train a multi-class classifier (i.e., the detector), and extend the free energy theory by making the following Remark \ref{remark1}.

\begin{remark}
    For AOSA tasks, the epistemic uncertainty (EU) of $x$ can be expressed through the free energy score on known classes minus that on the unknown class\footnote{$E_{unk}\left ( x \right )$ is determined by label-wise free energy (see \cref{sec:lwfe}).},
    \begin{equation}
    \label{eq3}
    \begin{aligned}
        &EU\left ( x  \right ) =E_{kno}\left ( x \right ) -E_{unk}\left ( x \right ) \\&=-\log_{}{\sum_{c=1}^{C}e^{-E\left ( x,c \right ) } }+\log_{}{\left ( 1+e^{-E\left ( x,C+1 \right ) } \right ) } .
    \end{aligned}
    \end{equation}
    As such, for two given data points $x_1$ and $x_2$, the inequality $EU\left ( x_1 \right )>EU\left ( x_2 \right )$ implies that $x_1$ is occurring from a denser region w.r.t. the unknown class compared to $x_2$.
\label{remark1}
\end{remark}

Since labeled data in AL is often quite limited, relying solely on the detector's predictions to assess the epistemic uncertainty of examples may not be sufficiently reliable. To obtain a more reliable measurement, we evaluate the epistemic uncertainty of examples in two ways: 1) a learning-based perspective that directly uses the detector's predictions, and 2) a data-driven perspective that relies on the similarity to labeled examples from different classes. To this end, we utilize the detector to extract features and emit $K$ arrows from each instance in the labeled data pool $\mathcal{D}_{L}$ to its $K$ nearest neighbors in the unlabeled data pool $\mathcal{D}_{U}$ based on cosine distance. For a data point $x_i^{U}$, its probability given class $y$ can be approximated as:
\begin{equation}
\label{eq4}
p\left ( x_i^{U}|y\right ) =\frac{\text{\# of Arrows}_{(x_j^L,y ) }       }{ \left | X^y \right | }  
\end{equation}
where $\text{\# of Arrows}_{(x_j^L,y ) } $ denotes the total number of arrows directed at $x_i^U$ from examples with label $y$, and $\left | X^y \right |$ represents the total number of examples with label $y$ in $\mathcal{D}_{L}$. If $x_i^{U}$ is in a region where examples with label $y$ densely exist, it is likely to receive more arrows, and vice versa.

Then, by virtue of Bayes' theorem, we define the data-centric class probability distribution of $x_i^{U}$ as
\begin{equation}
\label{eq5}
    p\left (y|x_i^{U}\right ) =\frac{p\left ( x_i^{U}|y\right )p\left ( y \right ) }{\sum_{c=1}^{C+1} p\left ( x_i^{U}|c\right )p\left ( c \right )} .
\end{equation}
Here, the prior $p\left ( y \right )$ is the probability of observing class $y$, and can be approximately determined by the sample count for each class in $\mathcal{D}_{L}$:
\begin{equation}
\label{eq6}
    p\left ( y \right )=\frac{\left | X^y \right | }{\sum_{c=1}^{C+1}\left | X^c \right | } .
\end{equation}
Based on Eqs. \eqref{eq1}, \eqref{eq4}, \eqref{eq5}, and \eqref{eq6}, we can obtain:
\begin{equation}
\label{eq7}
    p\left ( y|x_i^U \right )=\frac{\text{\# of Arrows}_{(x_j^L,y ) }  }{\sum_{c=1}^{C+1}\text{\# of Arrows}_{(x_j^L,c ) } } =\frac{e ^{-E\left ( x_i^U,y \right ) }}{e ^{-E\left ( x_i^U\right ) }}.
\end{equation}
As such, we can define the specific form of the energy function from a data-driven perspective to calculate the epistemic uncertainty of examples, as stated in Remark \ref{remark1}.

Here, for a given data point $x_i^U$, the uncertainty scores calculated in two different ways are denoted as $EU_L(x_i^U)$ and $EU_D(x_i^U)$. We first gather the uncertainty scores for all examples in $\mathcal{D}_{U}$ to form sets $\left \{ EU_L(x_1^U),\dots, EU_L(x_{\mathcal{N} _U}^U) \right \} $ and $\left \{ EU_D(x_1^U),\dots, EU_D(x_{\mathcal{N} _U}^U) \right \} $, and then fit two two-component Gaussian Mixture Models (GMMs) respectively to convert these scores into a probabilistic format. Suppose a tilde is added to denote the probabilistic format, we apply an element-wise product rule to obtain the final epistemic uncertainty score of $x_i^U$:
\begin{equation}
\label{eq8}
    \tilde{EU}(x_i^U)=\tilde{EU}_L(x_i^U)\odot \tilde{EU}_D(x_i^U).
\end{equation}

\textbf{Energy-based aleatoric uncertainty.} Aleatoric uncertainty arises through inherent noise in the data, that is, for the same instance $x$, different labels might be observed if multiple annotators label it independently. This means that aleatoric uncertainty can be defined based on the confusion between classes. As such, we further extend the free energy theory by making the following Remark \ref{remark2}.

\begin{remark}
\label{remark2}
    For AL tasks, the aleatoric uncertainty (AU) of $x$ can be expressed through the free energy scores on all classes minus that on secondary classes,
    \begin{equation}
    \begin{aligned}
        &AU(x)=E(x)-E_{\text{secondary classes}}(x)\\&=-\log_{}{\sum_{c=1}^{C}e^{-E\left ( x,c \right ) } }+\log_{}{\left [ \sum_{c=1}^{C}e^{-E\left ( x,c \right ) }- e^{-E\left ( x,y_{max} \right ) }\right ]  } ,
    \end{aligned}
    \end{equation}
    where $y_{max}$ is the most probable label for $x$. This implies that for two data points $x_1$ and $x_2$, if inequality $AU(x_1) > AU(x_2)$ holds, then $x_1$ occurs in a region closer to the decision boundary compared to $x_2$.
\end{remark}

Similarly, we gather the uncertainty scores for all examples in $\mathcal{D}_{U}$ to form set $\left \{ AU(x_1^U),\dots, AU(x_{\mathcal{N} _U}^U) \right \} $, and then fit a two-component GMM to convert these scores into a probabilistic format, i.e., $\left \{ \tilde{AU}(x_1^U),\dots, \tilde{AU}(x_{\mathcal{N} _U}^U) \right \} $. 

\textbf{Target-driven adaptive active sampling.} After having epistemic uncertainty scores and aleatoric uncertainty scores, we form the query set based on the strategy outlined in Figure \ref{fig:framework}. Specifically, in each active sampling round, we perform the querying in two rounds. In the first round, $kb$ examples with the lowest epistemic uncertainty scores are selected. Then, according to aleatoric uncertainty, we choose the top $b$ examples with the highest scores from the candidates obtained in the first round to query their labels.

Obviously, the choice of the $k$ value is critical. If $k$ is too small, such as $k=1$, it maximizes the likelihood that the queried examples belong to the closed-set classes. However, since the candidate set size matches the query set size, aleatoric uncertainty cannot effectively contribute. On the contrary, if the $k$ value is too large, the closed-set condition of the candidate set cannot be ensured, rendering aleatoric uncertainty meaningless.
Meanwhile, the optimal $k$ value often varies for different datasets. To enhance the strategy's generalizability, we introduce the expected target precision for known class queries to drive the adaptive adjustment of the $k$ value. The relation between the two is as follows:
\begin{equation}
\label{k1}
    k_{t+1}=\begin{cases}
k_t+a  & \text{ if } rP-tP>z ,  \\
k_t-a  & \text{ if } tP-rP>z  , \\
k_t  & \text{ if } \left | tP-rP \right |\le z ,
\end{cases}
\end{equation}
where $k_t$ is the value of $k$ in the $t$-th AL round, $tP$ is the expected target precision, $rP$ is the real query precision calculated as $ rP=\frac{\left | X_{kno}^{query} \right |  }{\left | X^{query} \right | } $ after Oracle labeling, $a$ is the variation amplitude and $z$ is the triggering threshold. Notably, although the number of hyper-parameters increases, setting them becomes significantly easier, and they are less sensitive to dataset variations. 

\subsection{Detector and Target Classifier Training}

\textbf{Detector training.} All labeled examples from the known classes and unknown classes are jointly used to train a detector with $C+1$ classes. For a given data point $x_i$ with label $y_i$, let $p_i$ denote its one-hot label, i.e., $p_{ic}$ is set to 1 and the others to 0, and $q_i$ denote its probability vector predicted by the detector. On the one hand, we use the following cross-entropy loss to train the detector:
\begin{equation}
    \mathcal{L} _{ce}^{x_i}=-p_i\log_{}{q_i} =-\sum_{c=1}^{C+1} p_{ic}\log_{}{q_{ic}} .
\end{equation}
On the other hand, we propose a margin-based energy loss to ensure that, for examples from known classes, the free energy scores for the first $C$ classes are high, while for examples from unknown classes, the free energy score for the $(C+1)$-th class remains low, referring to Remark \ref{remark1}.
The energy loss is calculated by\footnote{Replacing $E_{kno}(x_i)$ with $EU(x_i)$ is also a suitable choice; however, we find that the performance difference between the two is minimal.}:
\begin{equation}
    \mathcal{L} _{energy}^{x_i}=\begin{cases}
\left ( \max\left ( 0, E_{kno}(x_i)-m_{kno}  \right )  \right )^2   \text{ if } x_i \in \mathcal{D}_L^{kno},  \\
\left ( \max\left ( 0, m_{unk} - E_{kno}(x_i)  \right )  \right )^2  \text{if } x_i \in \mathcal{D}_L^{unk},
\end{cases}
\end{equation}
where $m_{kno}$ and $m_{unk}$ are the margins for known classes and unknown classes, respectively.

Thus, the total loss for training the detector is:
\begin{equation}
\label{l_detector}
     \mathcal{L} _{detector}^{x_i}  = \mathcal{L} _{ce}^{x_i} + \lambda _{e}\mathcal{L} _{energy}^{x_i},
\end{equation}
where $\lambda _{e}$ is a hyper-parameter that balances the two losses.

\textbf{Target classifier training.} All labeled examples from the known classes are used to train the target classifier by minimizing the standard cross-entropy loss:
\begin{equation}
\label{l_classifier}
     \mathcal{L} _{classifier}^{x_i}  = -p_i\log_{}{q_i} =-\sum_{c=1}^{C} p_{ic}\log_{}{q_{ic}}.
\end{equation}

The pseudocode of EAOA is shown in \cref{sec:pesudocode}.

%% file: sec/4_experi.tex
\begin{figure*}[!h]
\centering
   \includegraphics[width=0.98\linewidth]{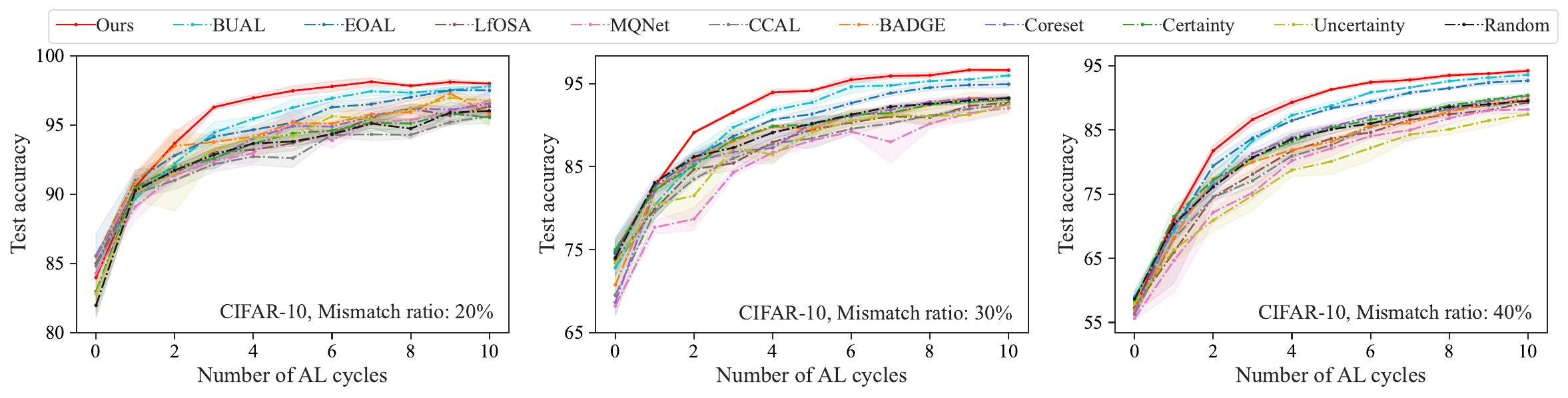}
   \includegraphics[width=0.98\linewidth]{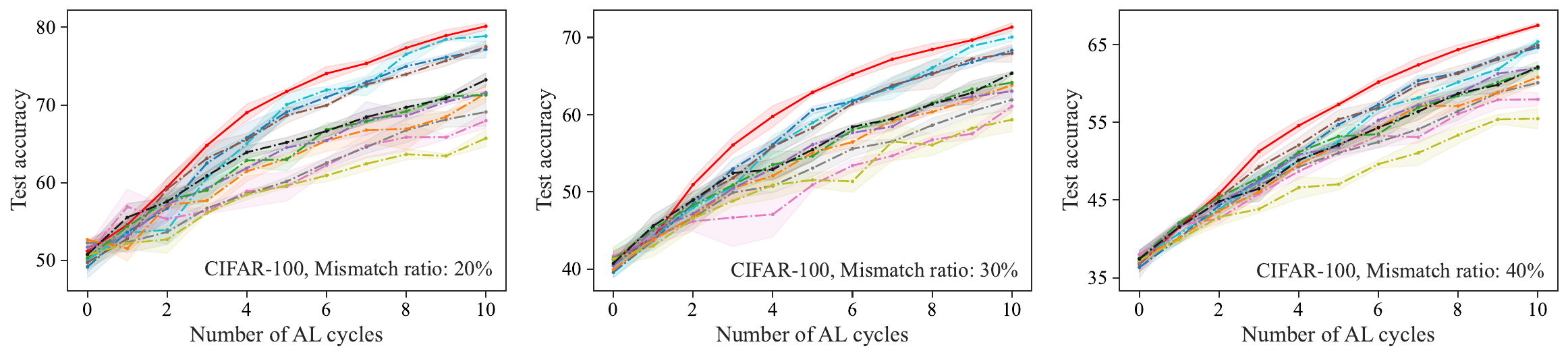}
   \includegraphics[width=0.98\linewidth]{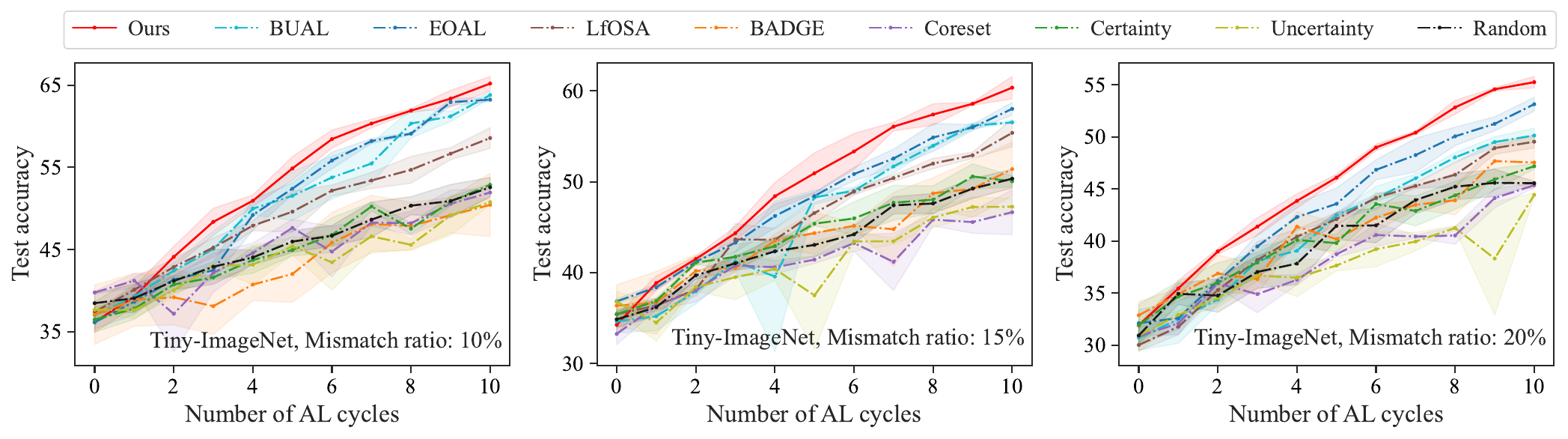}
\caption{Test accuracy comparison on CIFAR-10, CIFAR-100, and Tiny-ImageNet.}
	\label{fig:acc_cifar10}
\end{figure*}

\section{Experiments}
\label{sec:experiments}

\begin{figure*}[!h]
\centering
   \includegraphics[width=0.97\linewidth]{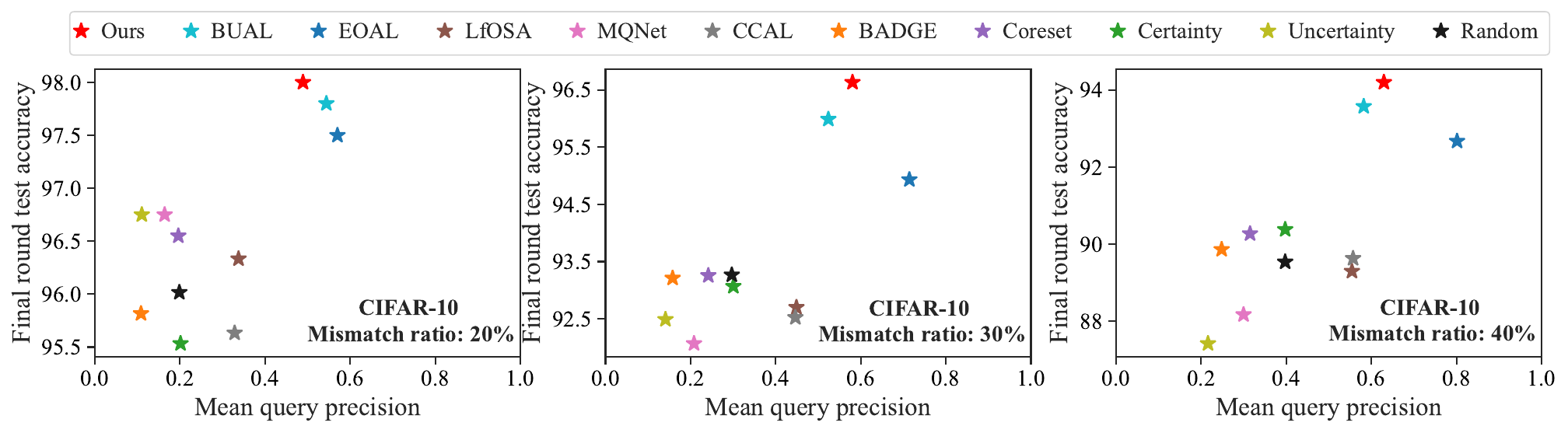}
   \includegraphics[width=0.97\linewidth]{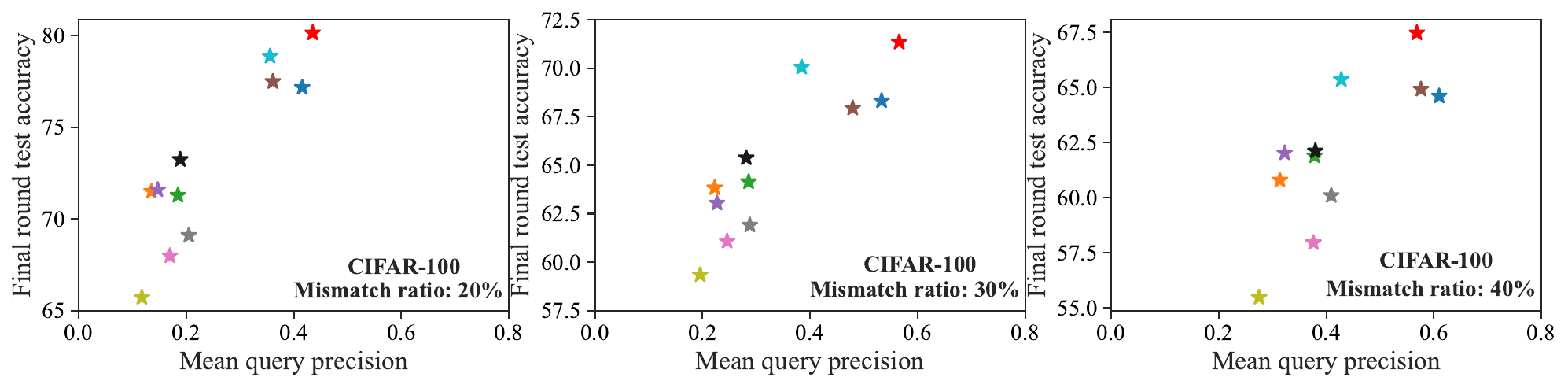}
   \includegraphics[width=0.97\linewidth]{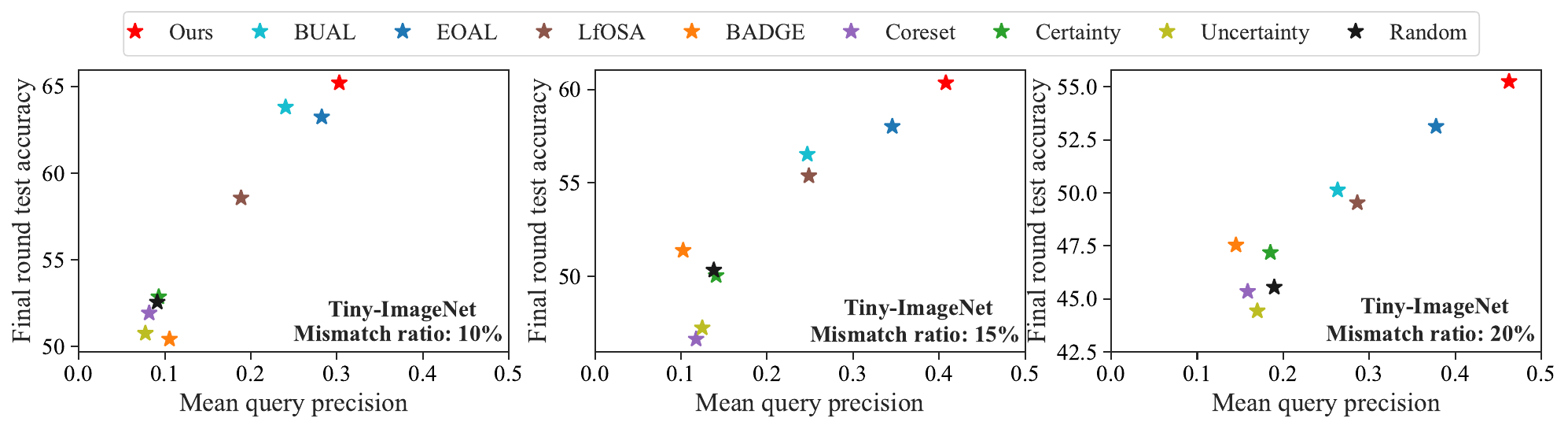}
\caption{Query precision comparison on CIFAR-10, CIFAR-100, and Tiny-ImageNet.}
	\label{fig:query_cifar10}
\end{figure*}


\subsection{Implementation Details}

\textbf{Datasets.} We validate the effectiveness of our method on three benchmark datasets: CIFAR-10 \cite{krizhevsky2009learning}, CIFAR-100 \cite{krizhevsky2009learning}, and Tiny-ImageNet \cite{yao2015tiny}, with category counts of 10, 100, and 200, respectively. To perform active open-set annotation (AOSA), we create their open-set versions by randomly selecting a subset of classes as known according to the specified mismatch ratio, while the remaining classes are treated as unknown. The mismatch ratio is defined as the proportion of known classes in the total number of classes. For CIFAR-10 and CIFAR-100, we set the mismatch ratios to 20\%, 30\%, and 40\%. For Tiny-ImageNet, we set the ratios to 10\%, 15\%, and 20\%, making it more challenging.

\textbf{Training details.} Initially, we randomly select 1\%, 8\%, and 8\% of known class examples from CIFAR-10, CIFAR-100, and Tiny-ImageNet, respectively, to construct the labeled dataset. The active learning (AL) process consists of 10 rounds, with 1,500 examples queried in each round.
For all experiments, we choose ResNet-18 \cite{he2016deep} as the base model and train it by SGD \cite{zinkevich2010parallelized} optimizer with momentum 0.9, weight decay 5e-4, and batch size 128 for 200 epochs. The initial learning rate is set to 0.01 and is reduced by a factor of 10 every 60 epochs. We repeat all experiments three times on GeForce RTX 3090 GPUs and record the average results for three random seeds  ($seed = 1, 2, 3$). We generally set the values of $K$, $tP$, $k_1$, $a$, $z$, $m_{kno}$, $m_{unk}$, and $\lambda_e$ to 250, 0.6, 5, 1, 0.05, -25, -7, and 0.01, respectively, and these values generalize well across datasets.


\textbf{Baselines.} We consider the following methods as baselines: Random, Uncertainty, Certainty, Coreset, BADGE, CCAL, MQNet, EOAL, and BUAL. Among them, EOAL and BUAL are currently state-of-the-art (SOTA). A detailed overview of these methods is provided in \cref{sec:baselines}.


\subsection{Performance Comparison} 

Figure \ref{fig:acc_cifar10} displays the test accuracy of various methods on CIFAR-10, CIFAR-100, and Tiny-ImageNet, varying with the number of AL rounds. Figure \ref{fig:query_cifar10} presents scatter plots of the average query precision across all rounds and the final round test accuracy for each method on CIFAR-10, CIFAR-100, and Tiny-ImageNet.
Here, the query precision refers to the proportion of queried known class examples to the total number of queried examples in each round.

As shown in Figure \ref{fig:acc_cifar10}, our method achieves optimal test accuracy across all datasets and mismatch ratios, and in most AL rounds, the curve of our method completely overlaps with those of other methods, demonstrating its superiority. In Figure \ref{fig:query_cifar10}, our method achieves optimal final round test accuracy and mean query precision in most scenarios, particularly on the challenging Tiny-ImageNet dataset, demonstrating its strong recognition capabilities.
Compared to the existing SOTA methods, BUAL and EOAL, our method ensures that queried examples exhibit low epistemic uncertainty while maintaining high aleatoric uncertainty. The significant test performance improvement over them validates the effectiveness of our proposed framework, suggesting that the examples selected by our method are more informative. All methods that employ a $(C+1)$-class detector to leverage labeled unknown class examples—BUAL, EOAL, LfOSA, and our method—exhibit significant performance advantages over the remaining methods, both in test performance and recognition capability. Although CCAL and MQNet consider both sample purity and informativeness, they fail to achieve an effective balance and adopt inadequate measurement metrics. Certainty shows similar recognition performance to Random, supporting that entropy, i.e., a measure of aleatoric uncertainty, is only meaningful in closed-set scenarios. Traditional AL methods, Uncertainty, Coreset, and BADGE, are significantly hindered by open-set examples, as these examples are prone to receive low-confidence predictions and exhibit diverse features, thus leading to poorer performance.

\begin{figure*}[!h]
	\centering
	
	\includegraphics[width=0.29\linewidth]{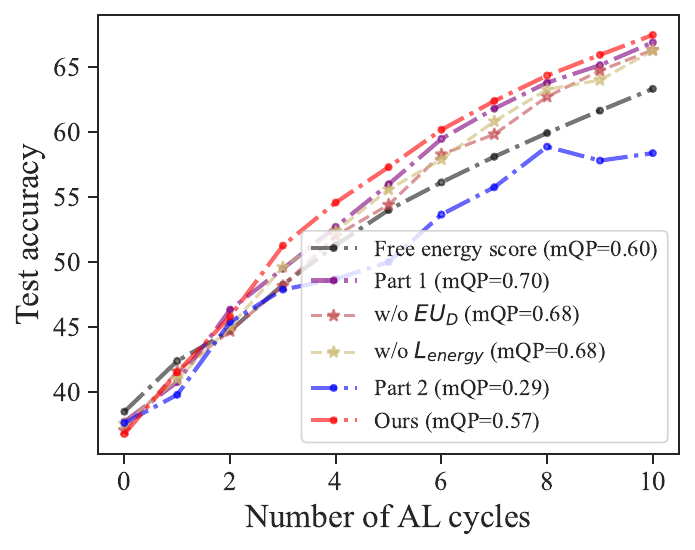}
	\includegraphics[width=0.29\linewidth]{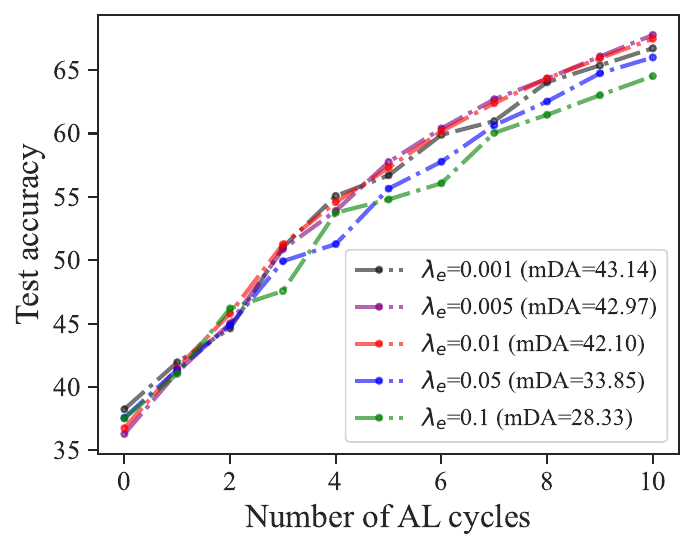}
        \includegraphics[width=0.41\linewidth]{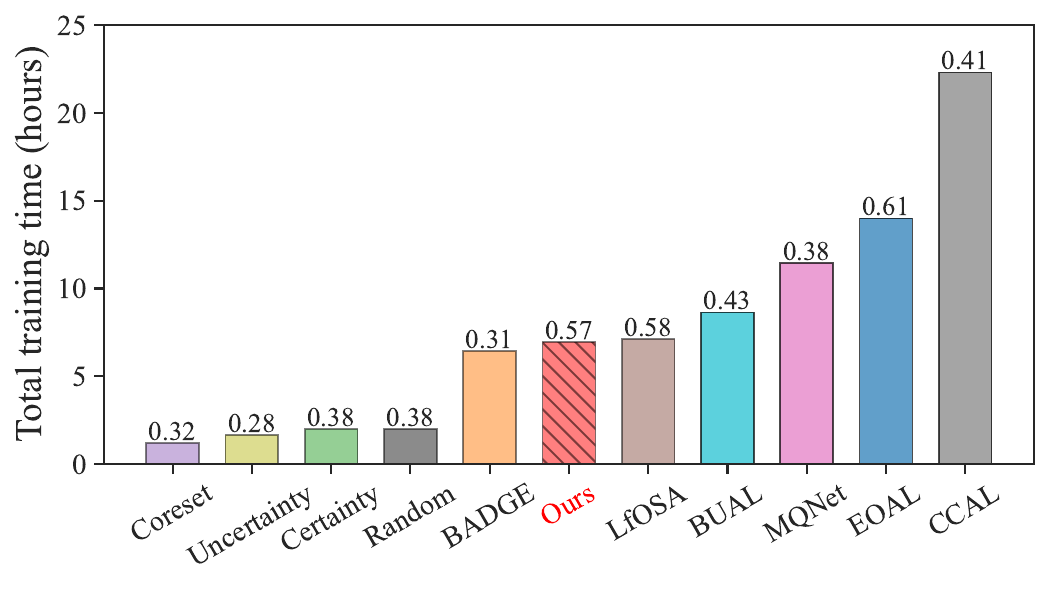}
	
	\caption{Ablation results on CIFAR-100 with a mismatch ratio of 40\%. 1) The \textbf{left} validates the effectiveness of each method component. ``Part 1" represents energy-based epistemic uncertainty and ``Part 2" indicates energy-based aleatoric uncertainty. ``mQP" denotes mean query precision across all AL rounds. 2) The \textbf{middle} assesses the sensitivity of the energy loss weight $\lambda_e$. ``mDA" denotes mean detector test accuracy across all AL rounds. 3) The \textbf{right} shows the runtime comparison. The numbers on the bar chart correspond to mQP scores.}
	\label{fig:three_aba}
\end{figure*}

\begin{figure}[!h]
  \centering
   \includegraphics[width=0.975\linewidth]{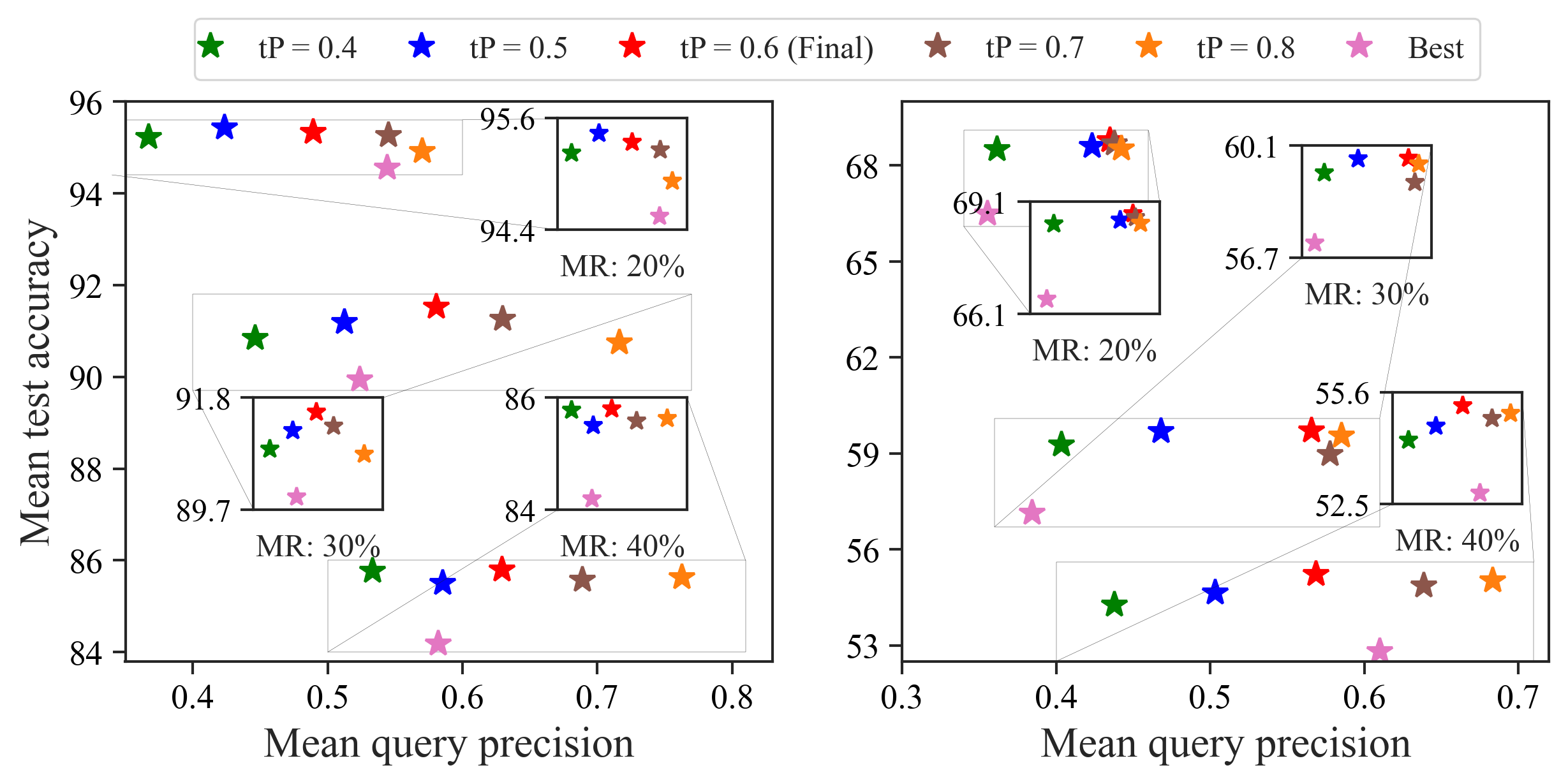}

   \caption{Ablation results for target precision $tP$ on CIFAR-10 (\textbf{Left}) and CIFAR-100 (\textbf{Right}). ``MR" denotes mismatch ratio. ``Best" indicates the top-performing method in the comparisons.}
   \label{fig:TR_aba}
\end{figure}

\subsection{Ablation Studies}

\textbf{Effect of each component.} Figure \ref{fig:three_aba} (left) shows the ablation results to validate the effectiveness of each component in our method. Our proposed energy-based epistemic uncertainty (``Part 1") utilizes information from labeled unknown class examples, leading to significant improvements in both recognition and test performance compared to free energy alone. The proposed energy-based aleatoric uncertainty (``Part 2") performs poorly on its own, as aleatoric uncertainty is only meaningful in closed-set scenarios. However, when combined with ``Part 1", it shows significant improvement, validating the superiority of the entire framework. Additionally, removing the data-driven epistemic uncertainty score or the margin-based energy loss leads to a decline in performance, confirming their necessity.

\textbf{Hyper-parameter sensitivity.} Figure \ref{fig:TR_aba} presents the ablation results for the hyperparameter target query precision $tP$ on CIFAR-10 and CIFAR-100, with values set to [0.4, 0.5, 0.6, 0.7, 0.8]. Overall, the fluctuations in test performance are minimal, and as $tP$ increases, the method's recognition performance improves. 
Figure \ref{fig:three_aba} (middle) displays the ablation results for the energy loss weight $\lambda_e$ on CIFAR-100 with a 40\% mismatch ratio, using values of [0.001, 0.005, 0.01, 0.05, 0.1]. A higher loss weight can impede model training, leading to reduced detector accuracy and reliability. Conversely, a lower loss weight fails to effectively separate the energy distributions of known and unknown class examples, adversely affecting the detector's recognition performance. 

The ablation results for additional hyper-parameters are provided in \cref{sec:aas}, which can demonstrate their ability to generalize effectively across different datasets.



\textbf{Runtime comparison.} Figure \ref{fig:three_aba} (right) presents the running times and mean query precision of various methods on CIFAR-100 with a mismatch ratio of 40\%. A higher mean query precision indicates that a larger number of examples are involved in training, often resulting in longer training times. Traditional AL methods, characterized by lower mean query precision and the training of a single model, generally have shorter training times. Notably, our method has the shortest runtime among all AOSA methods, achieving the highest test accuracy and maintaining a very high recognition rate.

	
	
	
	

%% file: sec/5_conclu.tex
\section{Conclusion}
\label{sec:conclusion}

In this paper, we demonstrate that focusing solely on either epistemic uncertainty (EU) or aleatoric uncertainty (AU) in open-world active learning scenarios does not yield satisfactory performance. We argue that the most informative examples should primarily belong to closed-set classes, exhibiting low EU scores, ensuring that the derived AU metric is meaningful, and secondarily, should show high AU scores. To achieve this, we propose EAOA, a novel framework for addressing the challenging active open-set annotation problem. In EAOA, both types of uncertainty are defined in the form of free energy: EU is evaluated by the detector from both learning-based and data-driven perspectives, while AU is measured by the target classifier through class confusion. Additionally, we introduce a margin-based energy loss to enhance the detector's ability to distinguish known from unknown classes and a target-driven strategy to adaptively adjust the size of the candidate set obtained in the first query stage. Extensive experimental results across various tasks demonstrate EAOA's superiority.

%% file: sec/X_suppl.tex
\clearpage
\setcounter{page}{1}
\maketitlesupplementary

\section{Uncertainty Quantification}
\label{sec:uq}

\begin{figure}[t]
  \centering
   \includegraphics[width=1\linewidth]{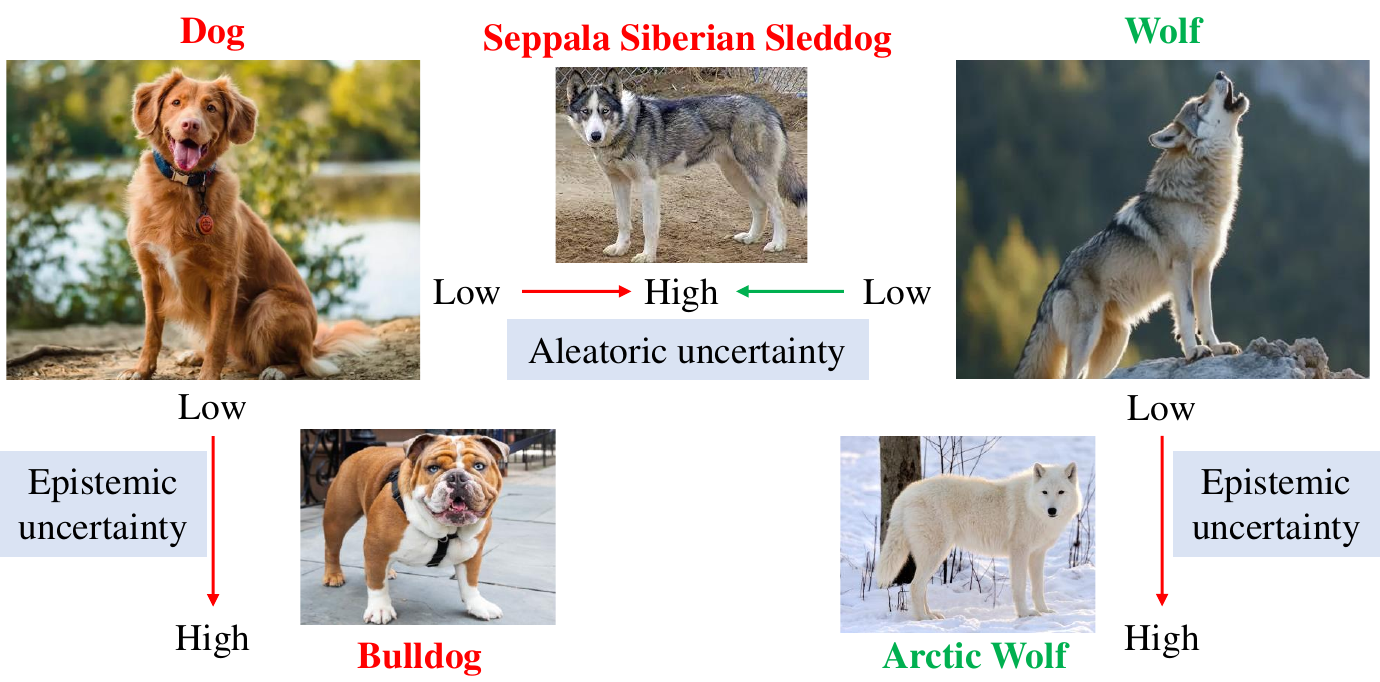}
   \caption{Intuitive examples of aleatoric and epistemic uncertainty in dog-wolf binary classification.}
   \label{fig:AU&EU}
\end{figure}

In deep learning, epistemic uncertainty and aleatoric uncertainty represent two distinct types of uncertainty, commonly used to describe the various sources of uncertainty in a model's predictions:
\begin{itemize}
    \item \textbf{Epistemic uncertainty:} 
    \begin{itemize}
    \item  This type of uncertainty arises from a model's lack of knowledge, often due to insufficient training data or the model's complexity. It reflects the model's incomplete or uncertain understanding of the task and can generally be reduced or eliminated with more data or a more effective model. 
    \item  For example, in a deep neural network, if there is little data available for certain classes or the model has not been trained sufficiently, the model may be highly uncertain in its predictions for certain examples.
    \item  This uncertainty is generally reducible, as it can be mitigated by adding more training data or improving the model architecture.
    \item  In Figure \ref{fig:AU&EU}, the ``Bulldog" and ``Arctic Wolf" exhibit significant feature differences from the ``Dog" and ``Wolf" in the training set, leading to higher epistemic uncertainty. After these examples are incorporated into model training, predictive performance on them improves, thereby reducing their epistemic uncertainty.
    \end{itemize}

    \item \textbf{Aleatoric uncertainty:} 
    \begin{itemize}
    \item  This type of uncertainty stems from inherent noise or variability in the data, i.e., the intrinsic randomness or uncontrollable factors within the data. 
    \item  For instance, in image classification, factors like feature confusion, lighting conditions, or object occlusion may lead to instability in the model's predictions.
    \item  This uncertainty is generally irreducible because it originates from the intrinsic properties of the data, not from issues with the model or training process.
    
    \item In Figure \ref{fig:AU&EU}, the ``Seppala Siberian Sleddog" resembles the ``Wolf" in appearance but belongs to the ``Dog" class, leading to higher aleatoric uncertainty. Due to feature confusion, incorporating these examples into model training may not substantially improve performance or reduce their aleatoric uncertainty.
    \end{itemize}
\end{itemize}

\section{Label-Wise Free Energy}
\label{sec:lwfe}

EBMs define the probability distribution in multi-label settings through the logits as:
\begin{equation}
\begin{aligned}
    p\left ( y_c|x \right ) &=\frac{e ^{-E\left ( x,y_c \right ) }}{\int_{y}e^{-E\left ( x,y_c \right ) } }=\frac{e ^{-E\left ( x,y_c \right ) }}{ e ^{-E\left ( x,y_c \right )}+e ^{-E\left ( x,-y_c \right )}}\\&=\frac{e^{-E\left ( x,y_c \right )+E\left ( x,-y_c \right )   } }{1+e^{-E\left ( x,y_c \right )+E\left ( x,-y_c \right )   } }=\frac{e^{f_{y_c}\left ( x \right ) }}{1+e^{f_{y_c}\left ( x \right ) } } \\&=\frac{e ^{-E\left ( x,y_c \right ) }}{e ^{-E\left ( x\right ) }}
\end{aligned}
\end{equation}
where $y_c=1$ indicates that instance $x$ belongs to the $c$-th class while $y_c=-1$ indicates not, $f_{y_c}\left ( x \right )$ denotes predicted logit of the model $f$ for instance $x$ regarding the $c$-th class, and $E_{y_c}\left ( x \right ) =-\log_{}{\left ( 1+e^{f_{y_c}\left ( x \right ) } \right ) } $ is the label-wise free energy for instance $x$ on class $y_c$.

\section{The Pseudocode of EAOA}
\label{sec:pesudocode}

The pseudocode of EAOA is summarized in Algorithm \ref{alg:1}.

\begin{algorithm}
    \caption{The EAOA algorithm}
    \label{alg:1}
    \textbf{Input:} Labeled data pool $\mathcal{D}_{L} = \mathcal{D}_{L}^{kno}\cup \mathcal{D}_{L}^{unk}$, unlabeled data pool $\mathcal{D}_{U}$, detector $f_{\theta_D}$, target classifier $f_{\theta_C}$, query budget $b$, dynamic factor $k_t$, and target precision $tP$. \\
    \textbf{Process: (The $t$-th AL round) }
    \begin{algorithmic}[1]
        \STATE \textit{\# Detector training}
        \STATE Update $\theta_D$ by minimizing $\mathcal{L} _{detector}$ in Eq. \eqref{l_detector} using all labeled examples from $\mathcal{D}_{L}$.

        \STATE \textit{\# Epistemic uncertainty estimating}
        \STATE Extract logit outputs and features from $f_{\theta_D}$ for examples in $\mathcal{D}_{L}$ and $\mathcal{D}_{U}$, respectively.
        \STATE Based on model outputs, estimate the learning-based epistemic uncertainty score for each example in $\mathcal{D}_{U}$ using Eq. \eqref{eq1} and Remark \ref{remark1}.
        \STATE Based on feature similarity, find K-nearest neighbors in $\mathcal{D}_{U}$ for each example in $\mathcal{D}_{L}$, and obtain reverse neighbors by class in $\mathcal{D}_{L}$ for each example in $\mathcal{D}_{U}$.
        \STATE Estimate data-centric epistemic uncertainty score for each example in $\mathcal{D}_{U}$ using Eq. \eqref{eq7} and Remark \ref{remark1}.
        \STATE For each example, combine the two scores into one final epistemic uncertainty score using GMM and Eq. \eqref{eq8}.
        
        \STATE \textit{\# Target classifier training}
        \STATE Update $\theta_C$ by minimizing $\mathcal{L} _{classifier}$ in Eq. \eqref{l_classifier} using all known class labeled examples from $\mathcal{D}_{L}^{kno}$.
        
        \STATE \textit{\# Aleatoric uncertainty estimating}
        \STATE Extract logit outputs from $f_{\theta_C}$ for examples in $\mathcal{D}_{U}$.
        \STATE Estimate aleatoric uncertainty score for each example in $\mathcal{D}_{U}$ using Remark \ref{remark2}.

        \STATE \textit{\# Active sampling}
        
        \STATE $k_tb$ examples with the lowest epistemic uncertainty scores are selected first to form a candidate query set.
        \STATE $b$ examples with the highest aleatoric uncertainty scores are then queried to form the final query set $X^{query}$.

        \STATE \textit{\# Oracle labeling}
        \STATE Query labels from Oracle and obtain $X_{kno}^{query}$, $X_{unk}^{query}$, and query precision $rP=\left | \frac{X_{kno}^{query}}{X^{query}}  \right | $.
        \STATE Update $k_t$ to $k_{t+1}$ using Eq. \eqref{k1} based on $tP - rP$.
        \STATE Update corresponding data pools: $\mathcal{D}_{U} = \mathcal{D}_{U} - X^{query}$, $\mathcal{D}_{L}^{kno}=\mathcal{D}_{L}^{kno}\cup X_{kno}^{query}$, and $\mathcal{D}_{L}^{unk}=\mathcal{D}_{L}^{unk}\cup X_{unk}^{query}$.
        
    \end{algorithmic}
    \textbf{Output:}  $\mathcal{D}_{L}$, $\mathcal{D}_{U}$, $\theta_D$, $\theta_C$, and $k_{t+1}$ for next round.
\end{algorithm}

\section{Comparing Methods}
\label{sec:baselines}

We consider the following AL methods as baselines: 
\begin{itemize}
\item Random, which selects instances at random; 

\item Uncertainty, which selects instances with the highest entropy of predictions; 

\item Certainty, which selects instances with the lowest entropy of predictions; 

\item Coreset, which uses the concept of core-set selection to choose diverse instances; 

\item BADGE, which selects instances by considering both uncertainty and diversity in the gradient via k-means++ clustering; 

\item CCAL, which employs contrastive learning to extract the semantic and distinctive scores of examples for instance querying; 

\item MQNet, which balances the purity score and informativeness score to select instances through meta-learning; 

\item LfOSA, which selects instances based on the maximum activation value produced by the $(C+1)$-class detector; 

\item EOAL, which queries instances by calculating the entropy of examples in both known and unknown classes; 

\item BUAL, which queries instances by adaptively combining the uncertainty obtained from positive and negative classifiers trained in different ways. 
\end{itemize}
Among these methods, EOAL and BUAL are currently state-of-the-art.

\section{Additional Ablation Studies}
\label{sec:aas}

\begin{figure}[!h]
  \centering
   \includegraphics[width=1\linewidth]{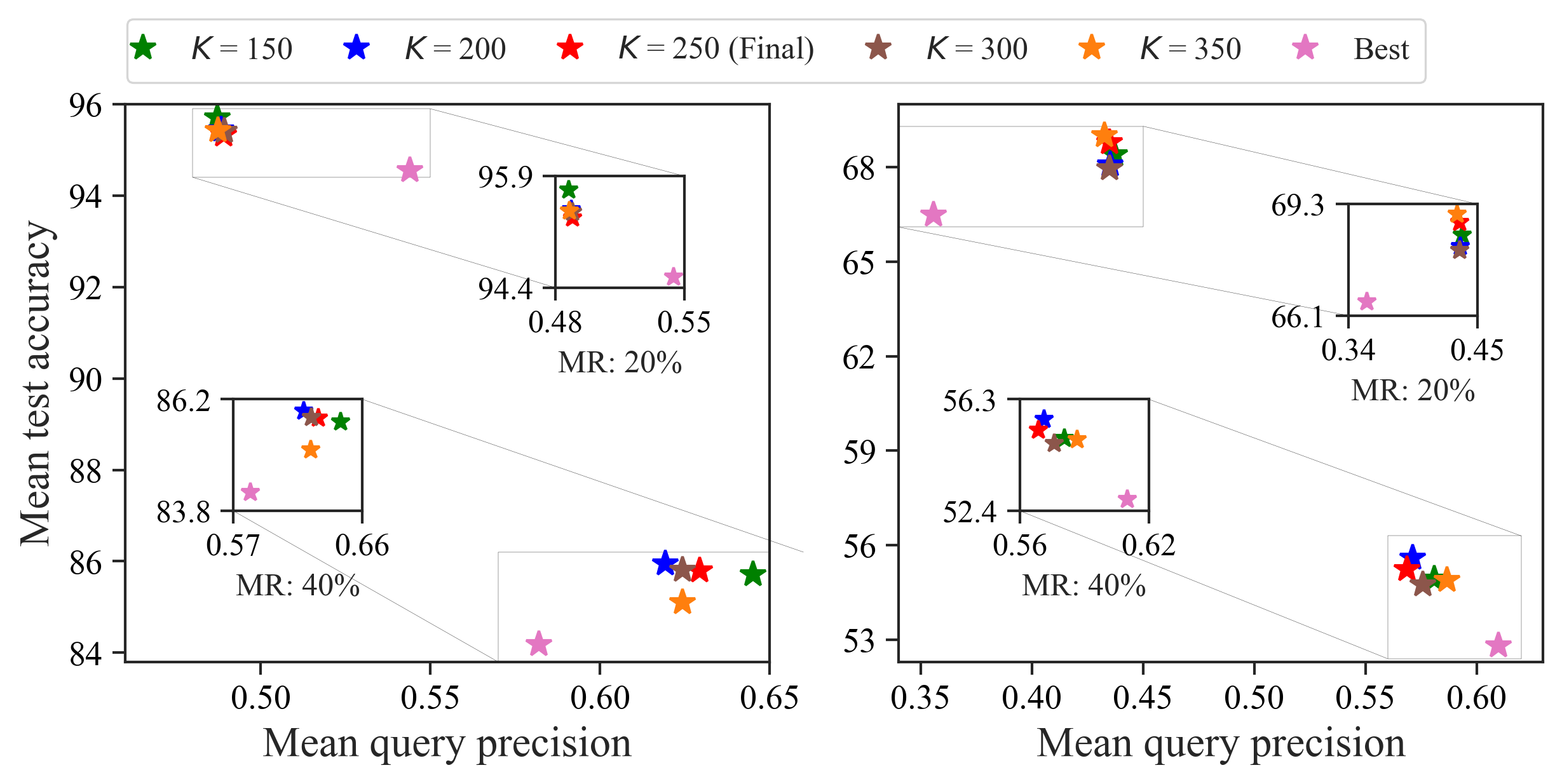}

   \caption{Ablation results for $K$ in reverse k-NN on CIFAR-10 (\textbf{Left}) and CIFAR-100 (\textbf{Right}). ``MR" denotes mismatch ratio. ``Best" indicates the top-performing method in the comparisons.}
   \label{fig:TR_aba_k}
\end{figure}

\begin{figure}[!h]
  \centering
   \includegraphics[width=1\linewidth]{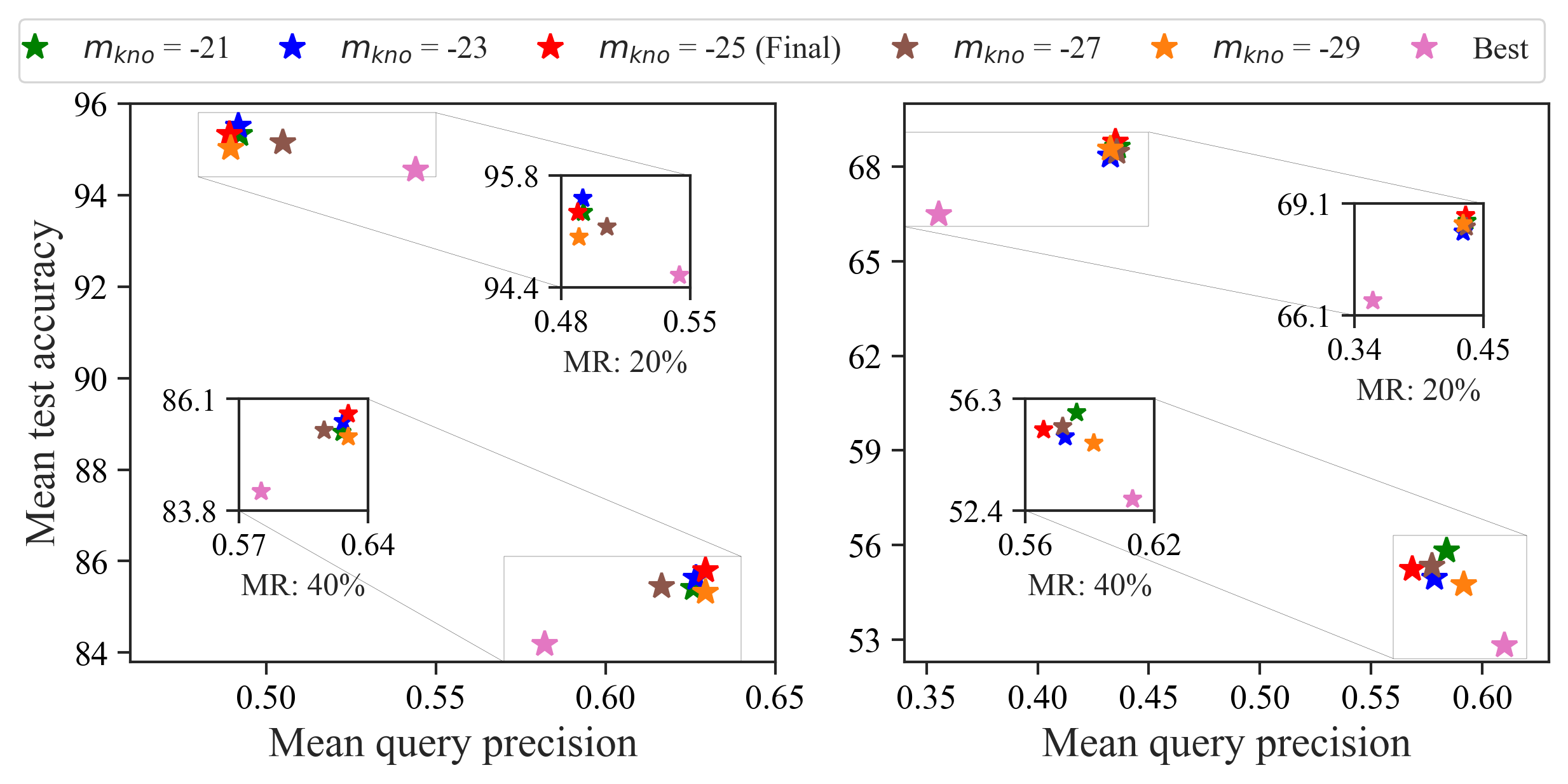}
   \includegraphics[width=1\linewidth]{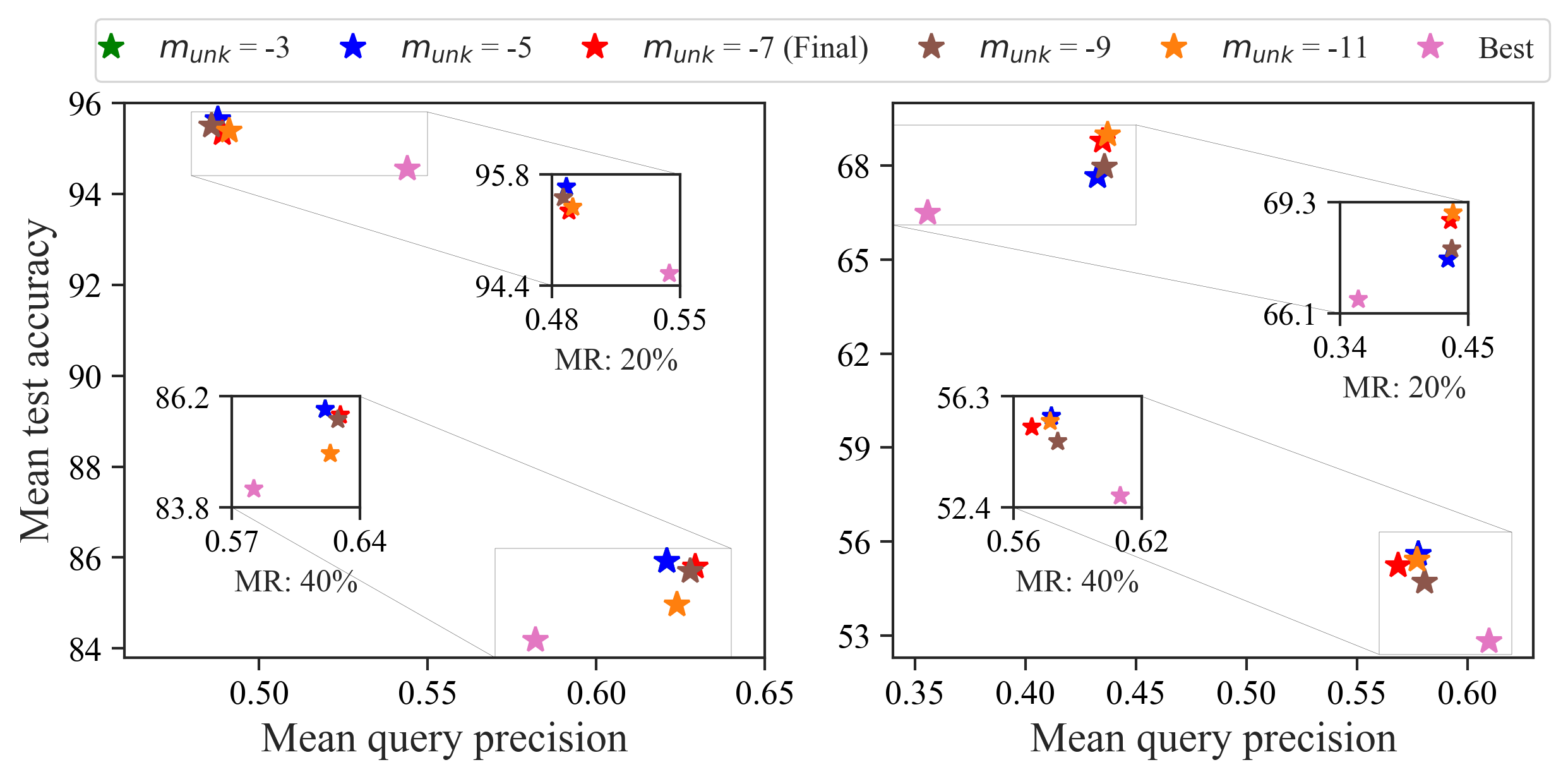}

   \caption{Ablation results for $m_{kno}$ and $m_{unk}$ in margin-based energy loss on CIFAR-10 (\textbf{Left}) and CIFAR-100 (\textbf{Right}). }
   \label{fig:TR_aba_m_kno_unk}
\end{figure}

Figure \ref{fig:TR_aba_k} illustrates the effect of the hyperparameter $K$ in reverse k-NN on EAOA's performance, with values set to [150, 200, 250, 300, 350]. 
Figure \ref{fig:TR_aba_m_kno_unk} presents the influence of the known class margin $m_{kno}$ and the unknown class margin $m_{unk}$ in margin-based energy loss $\mathcal{L} _{energy}$ on EAOA's performance, with values set to [-29, -27, -25, -23, -21] for $m_{kno}$ and [-11, -9, -7, -5, -3] for $m_{unk}$.
While the optimal value of $K$, $m_{kno}$, and $m_{unk}$ differ across different settings, their overall performance remains relatively stable compared to the top-performing method in the comparisons, with $K=250$, $m_{kno}=-25$, and $m_{unk}=-7$ consistently achieving strong results.


\begin{figure}[!h]
  \centering
   \includegraphics[width=1\linewidth]{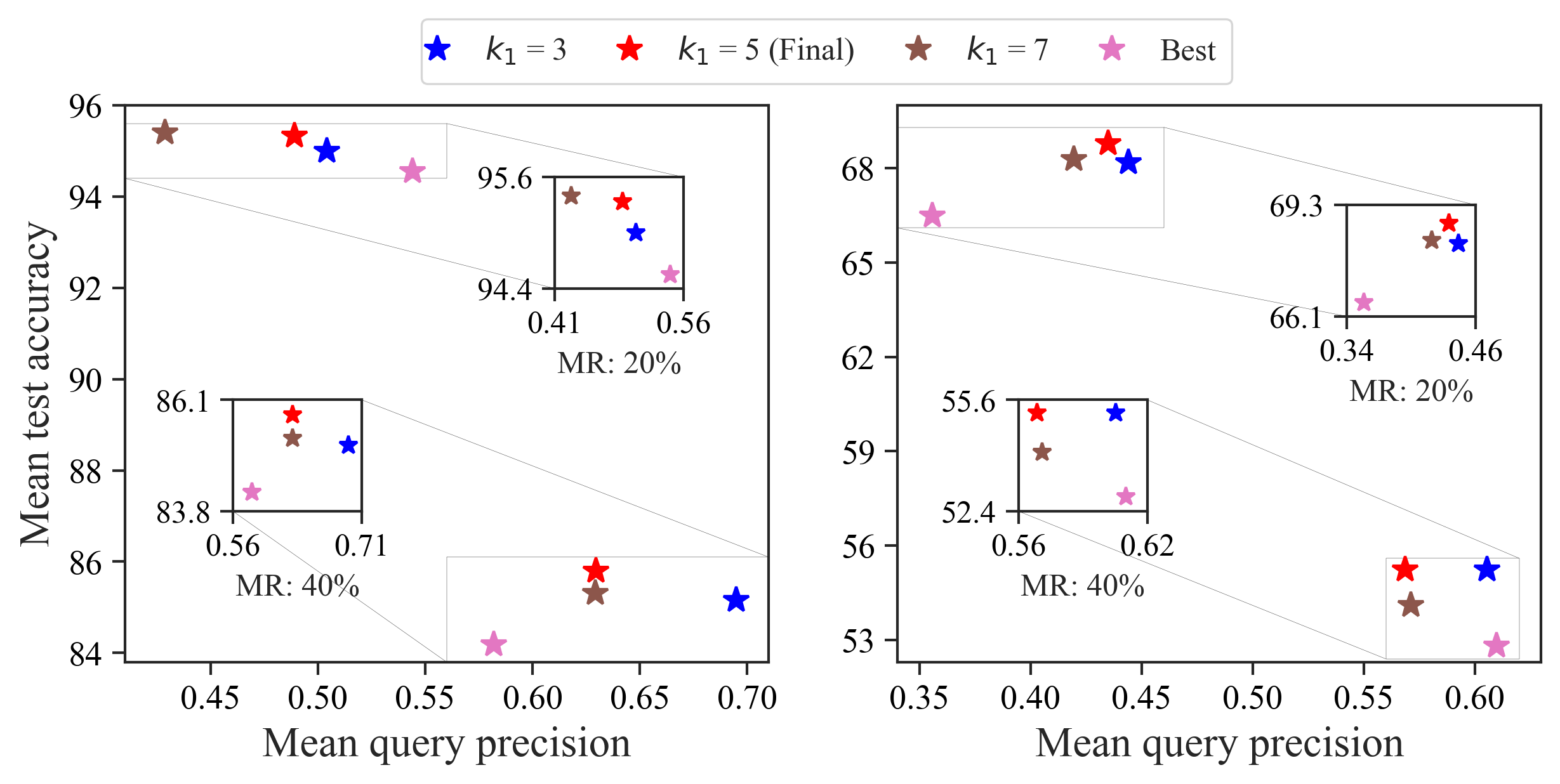}
   \includegraphics[width=1\linewidth]{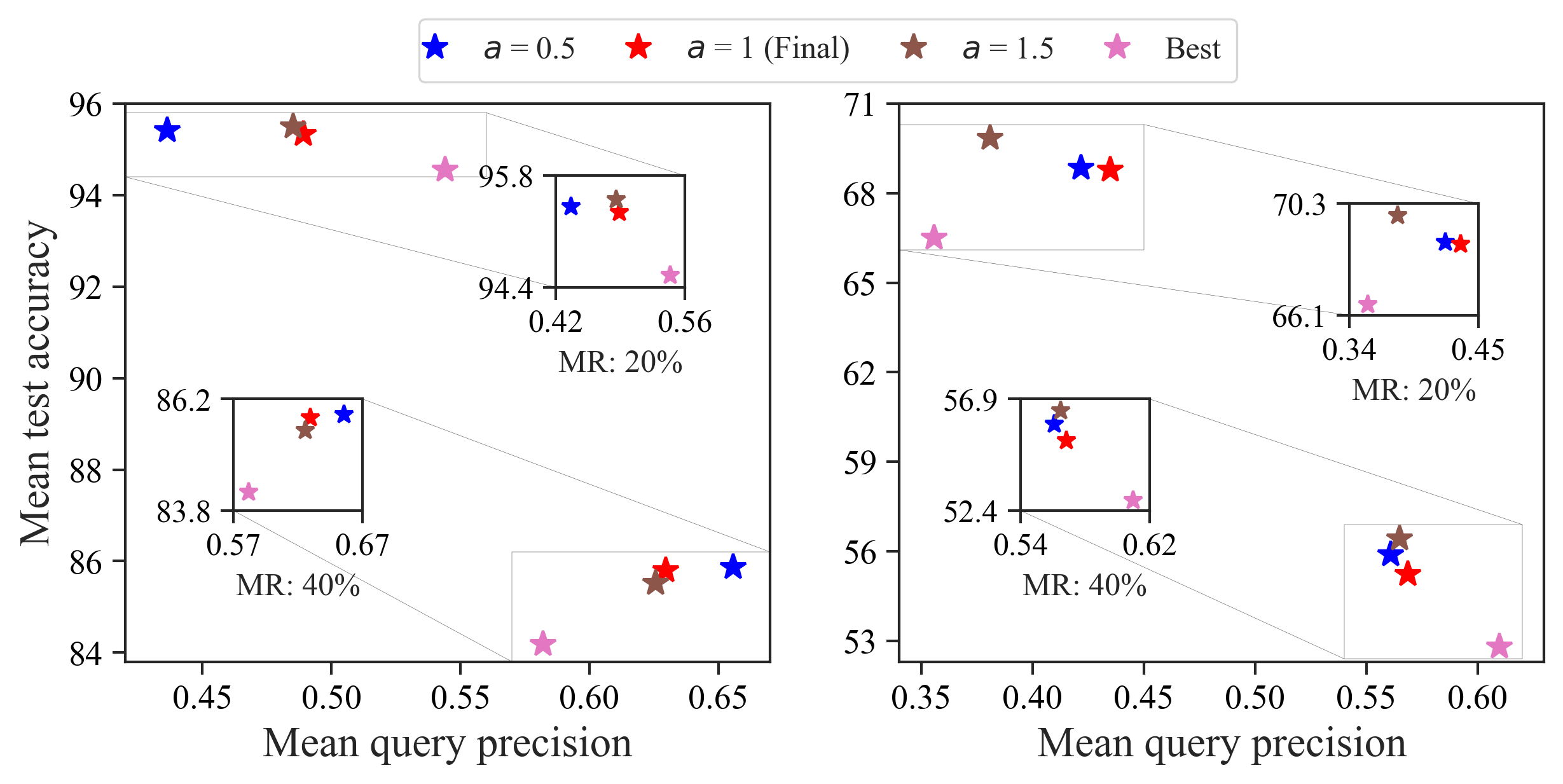}
   \includegraphics[width=1\linewidth]{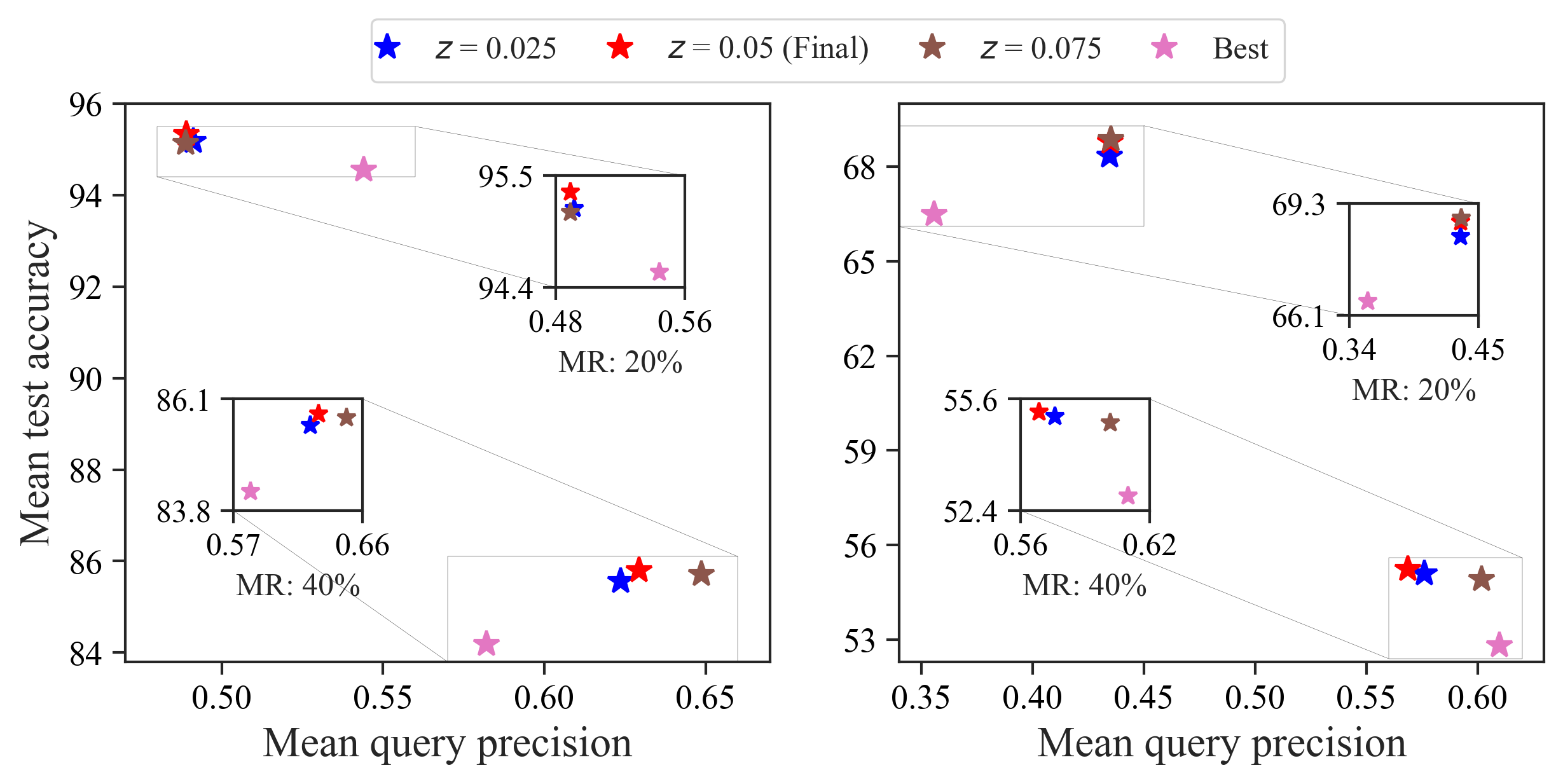}

   \caption{Ablation results for $k_1$, $a$ and $z$ in target-driven adaptive sampling strategy on CIFAR-10 (\textbf{Left}) and CIFAR-100 (\textbf{Right}).}
   \label{fig:TR_aba_k1}
\end{figure}

Figure \ref{fig:TR_aba_k1} shows the impact of initial round $k_1$, variation amplitude $a$, and triggering threshold $z$ in Eq. \ref{k1} on EAOA's performance, with values set to [-3, -5, -7] for $k_1$, [0.5, 1, 1.5] for $a$, and [0.025, 0.05, 0.075] for $z$. An excessively large $k_1$ value may lead to initial rounds that prioritize aleatoric uncertainty, beneficial for lower mismatch ratios. Conversely, a small $k_1$ value emphasizes epistemic uncertainty, making it suitable for higher mismatch ratios. Here, $k_1=5$ consistently delivers strong performance across various datasets. In practical applications, prior knowledge about the dataset can be used to further adjust its value. For hyperparameters $a$ and $z$, their values are simply set to 1 and 0.05 (ensuring no adjustments are triggered when the difference between target and actual query precision is within $\pm $0.05, or a range of 0.1) respectively to simplify parameter tuning. Although the parameter selection here is intuitive, the results in Figure \ref{fig:TR_aba_k1} confirm its suitability.

%
%
%

